\pdfoutput=1
\documentclass[11pt]{article}
\usepackage{acl}
\usepackage[english]{babel}
\usepackage[utf8]{inputenc}
\usepackage{times}
\usepackage{latexsym}
\usepackage[T1]{fontenc}
\usepackage{microtype}
\usepackage{inconsolata}
\usepackage[capitalize,noabbrev]{cleveref}
\usepackage{graphicx}
\usepackage{booktabs}
\usepackage{xspace}

\graphicspath{{img/}}

\setkeys{Gin}{width=\linewidth}

\makeatletter

\g@addto@macro\@floatboxreset\centering

\DeclareRobustCommand\onedot{\futurelet\@let@token\@onedot}
\def\@onedot{\ifx\@let@token.\else.\null\fi\xspace}

\def\eg{e.g\onedot} 
\def\ie{i.e\onedot}

\makeatother

\newcommand*\rot{\rotatebox{90}}

\title{CLoVe: Encoding \underline{C}ompositional \underline{L}anguage in \\ Contrastive \underline{V}ision-Language Models}

\author{Santiago Castro\thanks{Work conducted as an intern at Netflix.}\(^1\) \quad Amir Ziai\(^2\) \quad Avneesh Saluja\(^2\) \quad Zhuoning Yuan\(^2\) \quad Rada Mihalcea\(^1\) \\
\(^1\)University of Michigan -- Ann Arbor, \(^2\)Netflix \\
\texttt{sacastro@umich.edu}}

\begin{document}

\maketitle

\begin{abstract}
Recent years have witnessed a significant increase in the performance of Vision and Language tasks.
Foundational Vision-Language Models (VLMs), such as CLIP, have been leveraged in multiple settings and demonstrated remarkable performance across several tasks.
Such models excel at object-centric recognition yet learn text representations that seem invariant to word order, failing to compose known concepts in novel ways.
However, no evidence exists that any VLM, including large-scale single-stream models such as GPT-4V, identifies compositions successfully.
In this paper, we introduce a framework to significantly improve the ability of existing models to encode compositional language, with over 10\% absolute improvement on compositionality benchmarks, while maintaining or improving the performance on standard object-recognition and retrieval benchmarks.
Our code and pre-trained models are publicly available at \url{https://github.com/netflix/clove}.
\end{abstract}

\section{Introduction}

There has been a significant increase in the performance of Vision and Language tasks over the last few years \cite{clip,align,stable_diffusion,flamingo,idefics}.
Vision-Language Models (VLMs), such as CLIP~\cite{clip}, have been leveraged in multiple settings, either directly or indirectly as foundational models, and demonstrated remarkable performance across several tasks \cite{foundational_models,dall-e,dall-e-2,stable_diffusion,fitclip,blip2}.

\begin{figure}
\includegraphics{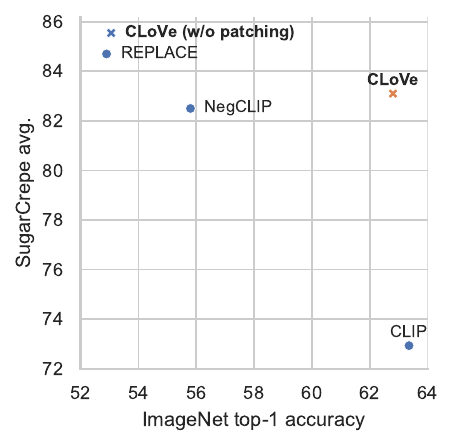}
\caption{Our proposed framework \textsc{CLoVe} significantly improves the compositionality performance (as measured by an average of SugarCrepe's seven fine-grained tasks) of pre-trained CLIP-like models while preserving their performance on other downstream tasks (as measured by ImageNet). Comparisons with more benchmarks are presented in \cref{tab:compositional-benchmark-results,tab:common-benchmark-results}. Baselines: REPLACE~\cite{sugarcrepe} and NegCLIP~\cite{aro}.}
\label{fig:main-results}
\end{figure}

Such models excel at object-centric recognition yet learn text representations that seem invariant to word order \cite{winoground,aro,castro-etal-2023-scalable}, failing to compose known concepts in novel ways~\cite{crepe,sugarcrepe}.
For example, as shown in \cref{fig:main-results}, CLIP has top performance on ImageNet tasks but falls behind on compositionality benchmarks.

Language compositionality is essential to recognizing more complex concepts in images or making text-to-image models successfully generate a novel scene with specific constraints~\cite{compositionality_in_visual_perception}.
For instance, in an image depicting \textit{``the woman shouts at the man,''} it is essential to recognize who is shouting at whom to understand the scene correctly.

Yet, no evidence exists that any VLM, including large-scale single-stream models such as GPT-4V~\cite{gpt4v}, identifies compositions successfully.
This assertion is supported by the fact that existing benchmarks that test compositionality continue to be an open challenge \cite{winoground,aro,crepe,sugarcrepe}.\footnote{See \cref{sec:related_work} for details.}

To address these limitations, previous work has introduced techniques to increase the compositional capabilities of pre-trained VLMs, such as NegCLIP~\cite{aro} and REPLACE~\cite{sugarcrepe}.
However, such methods come at a significant cost: they sacrifice the performance on more common object-centric recognition, as measured by ImageNet~\cite{imagenet}, EuroSAT~\cite{eurosat1,eurosat2}, and CIFAR100~\cite{cifar}.
For instance, as shown in \cref{fig:main-results}, NegCLIP showed an increase (compared to the pre-trained model) in its ability to address SugarCrepe~\cite{sugarcrepe} compositionality benchmark from 72.9\% to 82.5\% while, at the same time, its performance on ImageNet~\cite{imagenet} top-1 accuracy dropped from 63.4\% to 55.8\%.
Similarly, \citet{sugarcrepe} applied REPLACE to reach a high score of 84.7\% on SugarCrepe, but at the cost of a significant drop to 52.9\% on its ImageNet accuracy. 

In this paper, we introduce a framework to significantly improve the ability of existing two-tower models to encode compositional language while keeping the performance on more standard benchmarks, as shown in \cref{fig:main-results}.
Specifically, our contributions are as follows.
First, we show that \textbf{data curation} can significantly impact how a model can handle compositional knowledge.
Second, we confirm that training along with \textbf{hard negatives} can bring additional improvements.
Third, we show experimentally that \textbf{model patching} can be employed to preserve model performance on previous tasks.
Finally, we combine these ideas into a new framework called \textsc{CLoVe} and show that it can \textbf{significantly improve compositionality over a contrastively pre-trained VLM}. As a case study, we show how our framework can effectively improve CLIP's compositional abilities while maintaining the performance on other tasks.
Upon publication, we will provide checkpoints that others can use to substitute their CLIP-like model weights for a version with significantly better language composition abilities.

\begin{figure*}
\includegraphics{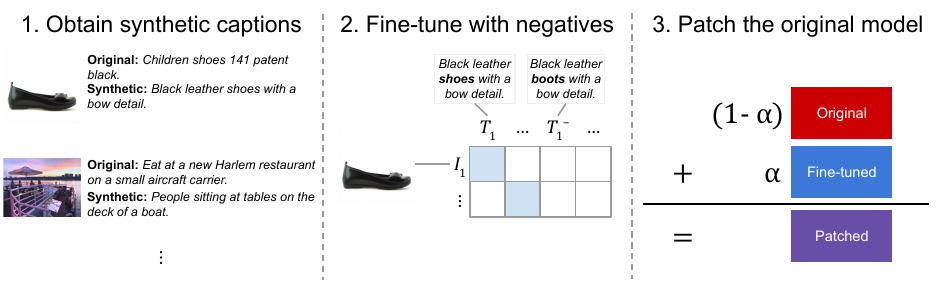}
\caption{Our \textsc{CLoVe} framework consists of three steps. First, obtain synthetic captions for a large image dataset. Second, fine-tune a pre-trained Contrastive VLM on it along with hard negative texts. Third, patch the original model with the fine-tuned one.}
\label{fig:method}
\end{figure*}

\section{Related Work}%
\label{sec:related_work}

\paragraph{Benchmarking Compositionality.}

Several frameworks have been proposed to measure model performance on language compositionality.
\citet{shekhar-etal-2017-foil} crafted a benchmark of foil image captions generated by changing a single word from the correct captions.
Models must identify if the image-caption pair correspond to each other, among other tasks.
Winoground~\cite{winoground} carefully built a high-quality dataset of 400 examples, each consisting of two images and two captions.
These two captions contain the exact word but in a different order following one of several strategies (\eg{}, swapping the subject and the object).
Each image must match the correct caption for the models to pass this test.
Models cannot simply rely on their ability to recognize concepts in images, as the elements repeat but are composed differently.

\citet{diwan-etal-2022-winoground} found that passing the Winoground benchmark successfully requires composition skills along with many others, such as commonsense reasoning and locating tiny objects.
\citet{aro} argued that Winoground is too small to draw statistically significant conclusions and built a benchmark called ARO consisting of examples with a single image, a correct caption, and multiple automatically generated incorrect captions.
CREPE~\cite{crepe} crafted a benchmark to measure compositionality in terms of systematicity and productivity.
It considers both seen and unseen compounds, among other phenomena.
SugarCrepe~\cite{sugarcrepe} is a recent benchmark that avoids ungrammatical and nonsensical negative captions while being large.
They showed it cannot be easily solved by computing the probability of the text captions without looking at the image.
Other benchmarks have also been created that consider compositionality as well as other phenomena, such as VALSE~\cite{parcalabescu-etal-2022-valse}, RareAct~\cite{rareact}, VL-Checklist~\cite{vl-checklist}, Cola~\cite{cola}, SVO-Probes~\cite{hendricks-nematzadeh-2021-probing}, and CLEVR~\cite{clevr}.

\paragraph{Methods to Improve Compositionality.}

Several works have shown that VLMs cannot recognize compositions successfully \cite{shekhar-etal-2017-foil,rareact,parcalabescu-etal-2022-valse,winoground,hendricks-nematzadeh-2021-probing,aro,castro-etal-2023-scalable,crepe}.
For this reason, NegCLIP~\cite{aro} was proposed to improve how CLIP~\cite{clip} composes concepts.
It consists of adding hard negative texts by taking the captions from the training batch and automatically generating sentences with the exact words but in a different order.
This approach makes the model distinguish between an image and the caption in the correct order compared to the exact words in an arbitrary order (as well as the other negative captions within the batch).
\citet{sugarcrepe} build upon NegCLIP and CREPE~\cite{crepe} and propose three ways to generate random negatives: REPLACE, SWAP, and NEGATE.
All these methods start from a Scene Graph representation of the sentence and operate over it.
REPLACE, which had the best overall results, performs single-atom replacements.
SWAP exchanges two atoms within the scene graph.
Finally, NEGATE introduces negation words (\ie{}, \emph{no} or \emph{not}).
We build upon NegCLIP~\cite{aro} and REPLACE~\cite{sugarcrepe} while we propose to use synthetically-generated captions to scale them up, as well as applying model patching~\cite{paint} to avoid catastrophic forgetting.
As far as we know, we introduce the first approach that significantly improves the composition skills of contrastively-trained models while preserving their zero-shot performance on other downstream tasks.

Cap and CapPa~\cite{cap} are two recently introduced models that employ captioning instead of contrastive learning (as in CLIP) to train VLMs.
\citet{cap} showed that they present an excellent performance on compositionality as measured by ARO~\cite{aro} and SugarCrepe~\cite{sugarcrepe}.
As these models rely on captioning and thus on computing the probability of the text given an image, they are inefficient for retrieval and classification.
For ARO, they showed that they can achieve high performance without looking at the image (they call it a ``blind decoder'').
For SugarCrepe, the authors did not compute this specific baseline.
Hence, we cannot infer the extent to which these models handle compositions successfully.
Our approach is different from them as it builds on top of contrastive two-tower models, which are efficient for retrieval and classification, and it does not rely on computing the probability of text, which is generally unimportant for such settings as all texts are equally likely (unlike in image captioning).

\section{\textsc{CLoVe}: A Framework to Increase Compositionality in Contrastive VLMs}

To address the compositionality limitations observed in previous models, we propose strategies to address the three main aspects of developing a contrastive VLM: data curation, contrastive learning, and model tuning.
We introduce \textsc{CLoVe}, a framework that leverages the strengths of an existing pre-trained contrastive VLM and enhances it with language composition skills.
\Cref{fig:method} shows an overview.

\textsc{CLoVe} includes the following steps, presented in more detail below:

\begin{description}

\vspace{-0.2cm} \item [3.1 Synthetic Captions.] Synthetic data generation can be effectively used to enlarge the training data. We use a large dataset with synthetic captions. 

\vspace{-0.2cm} \item [3.2 Hard Negatives.] Contrastive VLMs rely on the availability of negative training data. We add randomly generated hard text negatives to the dataset and train a fine-tuned model with increased compositionality capabilities. 

\vspace{-0.2cm} \item [3.3 Model Patching.] The pre-trained model and the fine-tuned model are combined through model patching. Patching allows us to keep the compositionality obtained with the fine-tuned model while recovering the pre-trained model performance on previously supported tasks.  
\end{description}

\subsection{Synthetic Captions}%
\label{subsec:synthetic}

Synthetic captions provide a great hybrid between the training dataset size and the quality of the captions.
We leverage LAION-COCO~\cite{laion-coco}, a 600-million dataset with images from the 2-billion-sized English subset of LAION-5B~\cite{laion5b} that were captioned with BLIP ViT-L/14~\cite{blip}, which was fine-tuned on COCO and filtered with two versions of OpenAI-pre-trained CLIP (\citealp{clip}; ViT-L/14 and RN50x64).
Even though the captions are limited in style (typically following the style of COCO captions), the LAION-COCO authors found that the synthetically generated captions have a similar quality to those written by humans.
We believe these captions focus more on describing visual information than the captions from its original dataset (LAION), based on multiple examples from this dataset.
See \cref{subsec:data_quality_exps} for an ablation of the training dataset.

\subsection{Hard Negatives}%
\label{subsec:training_with_hard_negatives}

Text hard negatives can enforce the model to better learn the meaning of each word, as they need to identify whether it relates to the image depending on how it is used in a caption.
\citet{aro} proposed NegCLIP, an extension of CLIP's training procedure that generates a hard negative text for each example in the batch by rearranging the image caption words.
These generated negatives are included within the negative test sets of the learning objective.
\citet{sugarcrepe} proposed an alternative called REPLACE and showed that the model can achieve better compositionality skills if such negatives are generated from carefully selected single-word replacements.
These replacements are performed on one of the entities, relations, or attributes obtained from first parsing the sentence as a scene graph, then selecting an alternative word from its antonyms or co-hyponyms by leveraging WordNet~\cite{wordnet}\footnote{More precisely, the method proposes to look for words that share a grand-co-hypernym.}.
These methods rely on high-quality captions.
Otherwise, the generated negatives will have changes that cannot be visually appreciated or will mostly be ungrammatical or nonsensical, and the model's downstream performance will be severely affected.
Take the following example from LAION that accompanies an image of a cardholder: \textit{``5x Orange Ball Wedding Party PLACE CARD HOLDER Table Name Memo Paper Note Clip.''}
If we apply REPLACE, supposing we can parse the sentence correctly, the word ``table'' could be replaced with ``bed''.
However, this would not make it a negative since the table is additional contextual information the caption included that cannot be visually appreciated.
Such a change will introduce more noise to the model's training process.

For this reason, these works have employed the COCO captions~\cite{coco,coco-captions} dataset.
COCO consists of images along with high-quality human-annotated captions that describe them.
Nevertheless, with 600,000 image-text pairs, COCO is at least three orders of magnitude smaller than the typically used image-text training datasets.
This issue limits learning and makes models overfit.
Additionally, COCO presents a limited number of objects and actions.
700 out of the 1000 object classes in ImageNet-1k are not present in COCO~\cite{noc}.
We propose combining these hard-negative techniques with a synthetic-caption dataset, such as LAION-COCO~\cite{laion-coco} (introduced in the previous subsection).

\subsection{Model Patching}%
\label{subsec:patching}

Model patching~\cite{paint} makes a fine-tuned model recover the performance on previously supported tasks while keeping the performance on the target task.
NegCLIP~\cite{aro} and REPLACE~\cite{sugarcrepe} fine-tune a model to significantly improve language compositional skills.
However, in exchange, they sacrifice the performance on general object recognition, as measured by their ImageNet performance.
For this reason, we propose applying one of such methods and subsequently employing model patching.
This procedure consists of performing a weight-space average between the pre-trained and the fine-tuned models.
Concretely, for each pre-trained model weight \(w^{PT}_i\) and fine-tuned model weight \(w^{FT}_i\), we compute their weighted average to obtain a new model weight \(w_i\):

\begin{equation}
w_i = (1 - \alpha) w^{PT}_i + \alpha w^{FT}_i
\end{equation}

In \cref{subsec:patching_exps}, we show that this approach helps the model gain compositionality properties while maintaining its object-recognition performance.

\section{Case Study on CLIP}

To demonstrate the effectiveness of our framework, we apply it to CLIP \cite{clip}, one of the most widely used contrastive VLMs.
Given that previous work has highlighted the tradeoff between compositionality abilities and model performance on previous standard tasks, we conduct evaluations both on challenging compositionality benchmarks as well as on standard benchmarks for object recognition and image-to-text and text-to-image retrieval.
To gain insights into the role played by the three main components of the \textsc{CLoVe} framework, we conduct three ablations studies to (1) determine the role of synthetic captions; (2) evaluate if employing hard negative texts during training improves the recognition performance of compositions; and (3) test the importance of patching the original model after training with hard negative texts.
Unless otherwise noted, all evaluations are zero-shot, meaning we do not perform in-domain fine-tuning on benchmark-specific training splits.

\begin{table*}
\small
\begin{tabular}{c|cccc|ccc|ccc|c}
 & \multicolumn{4}{c|}{ARO} & \multicolumn{3}{c|}{SugarCrepe} & \multicolumn{3}{c|}{SVO-Probes} \\
 & Attr\onedot{} & Rel\onedot{} & C-Ord\onedot{} & F-Ord\onedot{} & Repl\onedot{} & Swap & Add\onedot{} & Subj\onedot{} & Verbs & Obj\onedot{} & avg\onedot{} \\
\midrule
pre-trained & 63.5 & 59.8 & 47.7 & 59.9 & 80.1 & 62.3 & 72.8 & 84.0 & 79.3 & 87.8 & 69.7 \\
\midrule
NegCLIP & \underline{70.5} & \textbf{80.1} & 87.0 & 90.1 & 85.1 & \underline{75.3} & 85.9 & 90.9 & 84.7 & \underline{92.3} & \underline{84.2} \\
REPLACE & \textbf{71.2} & 72.9 & 80.1 & 86.7 & \underline{88.2} & 74.8 & \underline{89.5} & \textbf{92.0} & 84.6 & \underline{93.0} & 83.3 \\
\midrule
CLIP+\textsc{CLoVe} w/o patching & 69.0 & 77.4 & \textbf{91.7} & \textbf{93.6} & \textbf{88.6} & \textbf{76.1} & \textbf{90.5} & 88.2 & 83.7 & 91.6 & \textbf{85.0} \\
CLIP+\textsc{CLoVe} (\(\alpha = .6\)) & 69.7 & 72.7 & 86.6 & 92.1 & 87.0 & 74.6 & 85.8 & 90.5 & \textbf{86.4} & \textbf{93.3} & 83.9 \\
\end{tabular}
\caption{Zero-shot compositional evaluation results.}
\label{tab:compositional-benchmark-results}
\end{table*}

\begin{table*}
\small
\begin{tabular}{c|cccccccccc|c}
 & \rot{ImageNet} & \rot{Cars} & \rot{CIFAR10} & \rot{CIFAR100} & \rot{MNIST} & \rot{EuroSAT} & \rot{Flowers} & \rot{DTD} & \rot{UCF101} & \rot{HMDB51} & \rot{average} \\
\midrule
pre-trained & \textbf{63.4} & \textbf{59.7} & 89.8 & 64.2 & \textbf{48.9} & 50.5 & \textbf{66.6} & \textbf{44.4} & 69.3 & 44.3 & \underline{60.1} \\
\midrule
NegCLIP & 55.8 & 45.6 & 85.9 & 60.9 & 45.3 & 32.9 & 55.9 & 39.0 & 65.6 & 42.7 & 53.0 \\
REPLACE & 52.9 & 42.7 & 84.6 & 60.2 & 36.6 & 34.3 & 51.9 & 34.5 & 62.2 & 40.9 & 50.1 \\
\midrule
CLIP+\textsc{CLoVe} w/o patching & 53.1 & 48.7 & 88.5 & 62.0 & 40.4 & 46.9 & 43.2 & 36.3 & 62.3 & 41.0 & 52.2 \\
CLIP+\textsc{CLoVe} (\(\alpha = .6\)) & \underline{62.8} & 56.8 & \textbf{91.4} & \textbf{68.1} & \underline{48.7} & \textbf{57.4} & 61.1 & 41.2 & \textbf{70.4} & \textbf{46.0} & \textbf{60.4} \\
\end{tabular}
\caption{Zero-shot classification results.}
\label{tab:common-benchmark-results}
\end{table*}

\subsection{Experimental Setup}
\paragraph{Pre-trained Model.} 
Rather than starting from scratch, we aim to enhance the composition capabilities of an existing contrastive VLM.
This work uses CLIP (Contrastive Language-Image Pre-training;~\citealp{clip}), a pre-training method demonstrating impressive zero-shot performance on classification and retrieval tasks involving vision or language. 
It involves learning image and text representations in a joint space by leveraging large-scale weakly-supervised datasets.
These datasets contain image-text pairs with varying degrees of correspondence.
For each image, the model must learn the corresponding positive text from a set that includes this text and a random sample of \(N - 1\) other texts (negative samples) by employing the InfoNCE objective~\cite{infonce}.
Similarly, the model must identify which image corresponds to a given text.
CLIP is trained with mini-batch gradient descent, where this objective is applied to each pair in the \(N\)-sized batch, and the negatives are typically sourced from the rest of the batch.

\paragraph{Implementation Details.} 
Unless otherwise noted, the implementation details are the following.
We write our code on Python 3.10 using PyTorch~\cite{pytorch} v2.1, starting from \texttt{open\_clip}'s~\cite{openclip1,openclip2} codebase.
We run the experiments using the AdamW optimizer~\cite{adamw}, with a linear learning rate warmup for 2000 steps to 1e-6, later decayed with a cosine schedule~\cite{sgdr}.
We use a weight decay of 0.1.
Our initial pre-trained model is ViT-B-32 from OpenAI~\cite{clip}.
We train the models through one billion examples by randomly sampling with replacement from shards of up to \(10\,000\) samples, where the final size of each depends on the image availability at download time.
We successfully downloaded about 80\% of LAION-400M~\cite{laion}, 80\% of LAION-COCO~\cite{laion-coco}, and 60\% of COYO-700M~\cite{coyo} images.
The text captions are in English.
We employ one node with 8x A100 Nvidia GPUs and 96 CPU cores (\texttt{p4d.24xlarge} from AWS) for four days and a half.
The batch size is 256 per GPU.

\nocite{scipy,wolf-etal-2020-transformers,matplotlib,spacy,seaborn,pandas,numpy,nltk,tqdm,lhoest-etal-2021-datasets,jupyter,ftfy,hydra,ipython,timm,torchvision,parallel}

The choice of learning rate was based on multiple preliminary experiments to make sure it was not learning too slowly or that it was making the training loss go up.
The training steps and samples were selected to ensure we gave enough time for the method to learn and converge.
The choice of total batch size and compute budget was determined based on our availability compute and considering that CLIP-like methods need a large batch size.
All reported experiments are based on a single run since they are computationally expensive.

We re-implemented REPLACE~\cite{sugarcrepe} with the following changes and decisions, primarily because the code for this part is unavailable.
We skip employing BERT~\cite{devlin-etal-2019-bert} to filter the generated negatives
and instead proceeded to replace words based on the frequency of the new words, which is a first-order approximation of computing probabilities with a contextualized model.
For the replacements, given that the authors do not mention prepositions but we find them replaced in the provided data, we proceeded to replace prepositions.
For the replacement words, we try to respect the rest of the sentence by conjugating them (\eg{}, the person for the verbs, and the number for the nouns) and using a similar casing to the replaced word.
We used spaCy~\cite{spacy} v3.7.2 (the model \texttt{en\_core\_web\_sm}) and \texttt{pyinflect} v0.5.1.
We employed a different Scene Graph Parsing implementation, \texttt{SceneGraphParser} v0.1.0.
We avoid replacing a word with a potential synonym by looking at the synsets in common of their lemmas from WordNet~\cite{wordnet}, leveraging NLTK~\cite{nltk} v3.8.1.
We managed to reproduce the same numbers the original authors reported.
We will make our code publicly available to make it easy for anybody to reproduce and build on top of our results.

We set \(\alpha = 0.6\) for the model patching based on the ablation from \cref{subsec:patching_exps}.

\subsection{Using \textsc{CLoVe} to Bring Compositionality into CLIP}%
\label{subsec:main_exps}

We compare the CLIP model enhanced with our \textsc{Clove} framework against several baselines, as shown in \cref{fig:main-results}: CLIP+\textsc{Clove} leads to an average 10\% absolute improvement on the challenging compositionality benchmark SugarCrepe~\cite{sugarcrepe} when compared to a pre-trained CLIP model, all while maintaining its ImageNet performance within 1\%.
Additionally, we show that our model performs better than others on compositionality when we do not apply the model patching step.

In \cref{tab:compositional-benchmark-results}, we show a comparison of our enhanced CLIP+\textsc{Clove} model on others in three compositionality benchmarks: ARO~\cite{aro}, SugarCrepe~\cite{sugarcrepe} (over its three coarse-grained tasks), and SVO-Probes~\cite{hendricks-nematzadeh-2021-probing}.
Note that for SugarCrepe, we employ the macro-average to compute the coarse-grained task results like in \cite{cap} and unlike the original paper, since we are interested in measuring the global phenomena instead of giving importance to the task sample sizes.
See \cref{app_sec:sugarcrepe} for the performance on SugarCrepe for each fine-grained task.

Since a major concern in previous work when devising methods that increase model compositionality was the loss in performance on other tasks, we evaluate the CLIP+\textsc{Clove} model performance on object recognition and image-to-text and text-to-image retrieval tasks. 

In \cref{tab:common-benchmark-results}, we compare use the following object recognition benchmarks: ImageNet~\cite{imagenet}, Stanford Cars~\cite{cars}, CIFAR10~\cite{cifar}, CIFAR100~\cite{cifar}, MNIST~\cite{mnist}, EuroSAT~\cite{eurosat1,eurosat2}, Oxford Flowers 102~\cite{flowers}, Describable Textures (DTD)~\cite{dtd}, UCF101~\cite{ucf101}, and HMDB51~\cite{hmdb51}.
Following~\citet{clip}, we employ the top-1 accuracy metric, except for Oxford Flowers 102, where we use the mean per class.

\begin{table*}
\small
\begin{tabular}{c|cccc|cccc|c}
 & \multicolumn{4}{c|}{Text-to-Image/Video} & \multicolumn{4}{c|}{Image/Video-to-Text} & \\
 & \rot{CC3M} & \rot{DiDeMo} & \rot{MSR-VTT} & \rot{YC2} & \rot{CC3M} & \rot{DiDeMo} & \rot{MSR-VTT} & \rot{YC2} & avg\onedot{} \\
\midrule
pre-trained & 52.3 & 48.4 & 54.9 & 13.8 & 51.0 & 40.7 & 50.8 & 11.3 & 40.4 \\
\midrule
NegCLIP & 50.3 & 48.8 & 56.9 & 13.9 & 47.9 & 41.9 & 48.2 & 09.8 & 39.7 \\
REPLACE & 49.6 & \textbf{50.2} & 56.2 & 13.6 & 44.8 & 40.8 & 47.9 & 09.7 & 39.1 \\
\midrule
CLIP+\textsc{CLoVe} w/o patching & 47.3 & 35.0 & 53.1 & 11.4 & 43.4 & 37.8 & 42.7 & 08.0 & 34.8 \\
CLIP+\textsc{CLoVe} (\(\alpha = .6\)) & \textbf{58.7} & \underline{49.9} & \textbf{60.5} & \textbf{15.7} & \textbf{57.5} & \textbf{47.5} & \textbf{54.5} & \textbf{12.4} & \textbf{44.6} \\
\end{tabular}
\caption{Zero-shot retrieval results.}
\label{tab:common-benchmark-results-retrieval}
\end{table*}

In \cref{tab:common-benchmark-results-retrieval}, we present results on zero-shot text-to-image and image-to-text retrieval tasks.
The datasets used are: Conceptual Captions~\cite{sharma-etal-2018-conceptual} (CC3M), Distinct Describable Moments~\cite{didemo} (DiDeMo), MSR-VTT~\cite{msrvtt}, and YouCook2~\cite{youcook2} (YC2).
The results are presented by measuring Recall@5 -- the same metric used by~\citet{clip}.
Unlike in classification, our approach improves over the rest on average by at least 4\% (absolute).
We speculate this improvement comes from the fact that retrieval captions are longer and more complex than class labels, which allows us to appreciate our model's rich text representations.
We also believe using multiple prompts per class in classification tasks averages out the text representation noise from other models (see \cref{sec:classification_without_prompts} for an analysis of this).
Overall, we obtain better performance across all tasks and metrics using our \textsc{CLoVe} framework on CLIP, except for DiDeMo in text-to-image, whose performance is on par with REPLACE.

\subsection{Ablation Studies}

\paragraph{The Importance of Synthetic Captions.} 
\label{subsec:data_quality_exps}

\begin{table}
\small
\begin{tabular}{c|cccc}
Fine-tuning dataset & Attr\onedot{} & Rel\onedot{} & C-Ord\onedot{} & F-Ord\onedot{} \\
\midrule
pre-trained & 63.5 & 59.8 & 47.7 & 59.9 \\
\midrule
\multicolumn{5}{c}{\textit{Without hard negative texts}} \\
\midrule
COYO & 63.6 & 55.4 & 34.8 & 43.4 \\
LAION (L) & \underline{64.9} & 64.0 & 40.2 & 47.0 \\
COCO (C) & 62.5 & 61.6 & \textbf{73.8} & 39.8 \\
concat\onedot{} L \& C & \textbf{65.9} & 59.0 & 43.7 & 50.3 \\
sample unif\onedot{} L \& C & 64.6 & 55.7 & 59.8 & 29.7 \\
LAION-COCO & \underline{65.4} & \textbf{66.0} & 70.5 & \textbf{76.9} \\
\midrule
\multicolumn{5}{c}{\textit{With hard negative texts}} \\
\midrule
COYO & \underline{69.5} & 75.6 & 71.7 & 79.7 \\
LAION (L) & 67.9 & 72.6 & 78.3 & 85.4 \\
COCO (C) & \textbf{70.2} & 67.6 & \underline{90.9} & 74.5 \\
concat\onedot{} L \& C & \underline{70.1} & 76.2 & 83.4 & 88.6 \\
sample unif\onedot{} L \& C & \underline{69.9} & 71.6 & 82.7 & 60.8 \\
LAION-COCO & 69.0 & \textbf{77.4} & \textbf{91.7} & \textbf{93.6} \\
\end{tabular}
\caption{The zero-shot performance of fine-tuning CLIP with different datasets, with and without hard negative texts.}
\label{tab:ablation-dataset}
\end{table}

We hypothesize that training dataset quality is essential to model compositionality performance.
For example, in LAION~\cite{laion}, a dataset commonly used to train Contrastive VLMs, you can find examples that present excessive information that cannot be easily mapped to visual concepts depicted in any image, such as: \textit{``Platinum Dance Academy T-shirt. Orders must be placed by Friday, September 26th. Delivery approximately 2 weeks or less.''}

Datasets with high-quality annotations such as COCO~\cite{coco,coco-captions} can be used, but such datasets are typically small (less than a million samples). 
A hybrid approach, with high-quality data and a large dataset, can be obtained using synthetic captions, as described in \cref{subsec:synthetic}.
We are interested in comparing this dataset with LAION-400M or COCO directly, as well as two ways to combine the datasets:
a) concatenation and b) sampling with equal probability.\footnote{Note LAION-400M is about 700 times larger than COCO.}
Note that these strategies of combining LAION and COCO are completely different from the LAION-COCO dataset 
In addition, we consider COYO-700M~\cite{coyo}, a large-scale dataset constructed similarly to LAION-400M.

\Cref{tab:ablation-dataset} compares the performance of fine-tuning a pre-trained CLIP model on different datasets without employing negatives.
In this table and subsequent ones, the best results are in \textbf{bold}, and an \underline{underline} indicates results within 1\% of best.
LAION-COCO~\cite{laion-coco} presents the best results overall, with a large margin on ARO.
For this benchmark, it is the only presented dataset that significantly outperforms the pre-trained model.
In the case of the SugarCrepe benchmark, we observe that all datasets provide improvements over the pre-trained model.
Interestingly, \citet{dall-e-3} also found synthetic captions helpful for text-to-image generation models.
They show synthetic captions help such models generate images that align better with the input text.

\paragraph{The Importance of Hard Negatives.} %
\label{subsec:hard_negatives_exps}

\begin{table}
\scalebox{0.85}{
\begin{tabular}{c|cccc}
 & Attr\onedot{} & Rel\onedot{} & C-Ord\onedot{} & F-Ord\onedot{} \\
\midrule
pre-trained & 63.5 & 59.8 & 47.7 & 59.9 \\
\midrule
fine-tuned & 65.4 & 66.0 & 70.5 & 76.9 \\
+ negatives & \underline{69.0} & \textbf{77.4} & \textbf{91.7} & \textbf{93.6} \\
\midrule
+ negatives* & \textbf{69.4} & 75.4 & 77.5 & 86.1 \\
\end{tabular}
}
\caption{The importance of employing negatives to improve the zero-shot performance on recognizing compositions. *The last row shows the results of using half the batch size -- there are gains even when the total device memory is the same, given that employing negatives effectively doubles the batch size.}
\label{tab:ablation-negatives}
\end{table}

\begin{figure}[ht]
\includegraphics{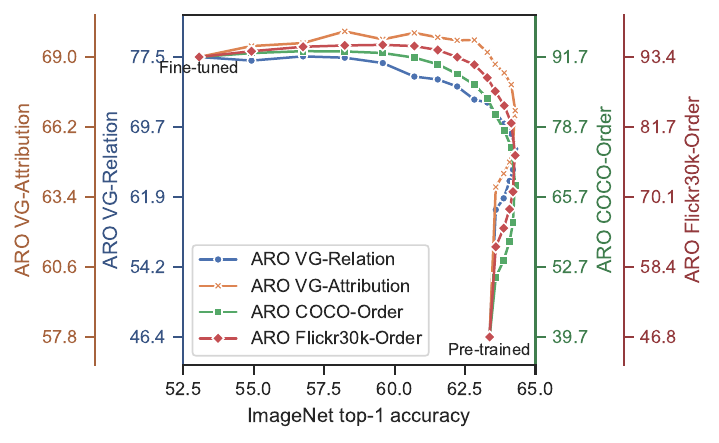}
\caption{The effect of applying model patching to both an object-centric benchmark (ImageNet,~\citealp{imagenet}; x-axis) and a compositionality benchmark (ARO,~\citealp{aro}; the four y-axes represent its four tasks), when varying the value of the weight in the average, \(\alpha\). The value of \(\alpha\) varies from 0 (the pre-trained model) to 1 (the fine-tuned model) in 0.05 increments, and the lines connect such points. We can obtain models with good zero-shot performance in ImageNet and compositionality when \(\alpha\) is around 0.4--0.7. Note the four y-axes were adjusted to make the pre-trained and fine-tuned model points match to focus on how the lines vary between them.}
\label{fig:ablation-patching-alpha}
\end{figure}

\citet{aro,sugarcrepe} showed that employing randomly generated text negatives as part of the training process can significantly improve the language compositionality skills of pre-trained models.
We apply REPLACE~\cite{sugarcrepe} to obtain randomly generated hard negative text along with the LAION-COCO dataset~\cite{laion-coco} and compare it to fine-tuning without negatives.
We present the results in \cref{tab:ablation-negatives}.
In this setting, we can observe that employing negatives improves performance over not using them, as measured by the ARO benchmark~\cite{aro} (its tasks are, in the order that we show them: VG-Attribution, VG-Relation, COCO-Order, and Flickr30k-Order).

\paragraph{The Importance of Model Patching.}%
\label{subsec:patching_exps}

Existing methods to improve CLIP's compositionality by employing negatives used by \citet{aro,sugarcrepe} do so by considerably hurting the model's performance on more standard object-centric benchmarks such as ImageNet~\cite{imagenet}.

\Cref{fig:ablation-patching-alpha} presents the effect of varying this value for both a compositionality benchmark and an object-centric one.
When \(\alpha\) is around 0.4--0.7, the model performs well on both.

\section{Conclusions}

In this paper, we introduced \textsc{CLoVe} -- a framework to considerably improve the compositionality of pre-trained Contrastive VLMs while preserving their performance on other tasks, unlike existing methods.
Our approach combines fine-tuning contrastive VLMs with hard negative texts by leveraging synthetically captioned images, as they can provide an excellent tradeoff between quality and quantity.
Subsequently, it patches the original model with the fine-tuned one to convey the best of two worlds -- compositional skills while maintaining the performance on other tasks.

We showed experimentally that \textsc{CLoVe} improves the performance of CLIP-like models on multiple benchmarks, both compositionality-related and non-compositionality-related.
We ablated the different components of our framework and showed their importance: data quality, the use of hard negatives in training, and model patching.

Our code and pre-trained models are publicly available at \url{https://github.com/netflix/clove}.
Our code will allow for an easy replacement of CLIP-like weights with the ones we provide, considerably boosting the language composition performance.

\section*{Limitations}

Our work is limited in the following ways.

Our approach does not solve the compositionality problem completely.
Its performance on the compositionality benchmarks still presents a gap regarding the human performance reported by the papers associated with each of the employed benchmarks.

Employing synthetic captions can introduce undesired noise.
Image captioners may sometimes hallucinate, introducing incorrect concepts or inaccurate descriptions of such objects.
This is especially true for quantities, such as when there are four horses in the scene, but the synthetic caption mentions three.
Future work can focus on methods to improve the synthetic caption quality.

We did not study the effect of the performance of the patched models on different demographics.
It could be the case that some demographics are misrepresented in some task performance (compositional or not) after the model has been patched.
Users should be careful about this aspect.

In this work, we focus on two-tower models because of their efficiency for classification and retrieval.
We leave the study of single-tower models for future work.

\section*{Acknowledgements}

We thank Pablo Delgado and Netflix's training platform team for their help with using Netflix's computational resources.
We thank Muhammad Khalifa, Oana Ignat, Andrew Lee, and the Language and Information Technologies group at the University of Michigan for multiple insightful discussions.
This material is partly based on work supported by the Automotive Research Center (``ARC'').
Any opinions, findings, conclusions, or recommendations expressed in this material are those of the authors and do not necessarily reflect the views of ARC or any other related entity.

\bibliography{anthology,custom}

\appendix

\section{SugarCrepe Fine-Grained Performance}%
\label{app_sec:sugarcrepe}

In \cref{tab:sugarcrepe-results}, we show SugarCrepe's fine-grained task results.

\begin{table*}
\small
\begin{tabular}{c|ccc|c|cc|c|cc|c|cc}
 & \multicolumn{4}{c|}{Replacement} & \multicolumn{3}{c|}{Swap} & \multicolumn{3}{c|}{Addition} & & \\
 & Obj\onedot{} & Att\onedot{} & Rel\onedot{} & avg\onedot{} & Obj\onedot{} & Att\onedot{} & avg\onedot{} & Obj\onedot{} & Att\onedot{} & avg\onedot{} & \rot{task avg\onedot{}} & \rot{avg\onedot{}} \\
\midrule
pre-trained & 90.8 & 80.2 & 69.1 & 80.1 & 61.0 & 63.8 & 62.3 & 77.1 & 68.5 & 72.8 & 71.7 & 72.9 \\
\midrule
NegCLIP & 92.6 & 85.9 & 76.8 & 85.1 & \textbf{75.6} & 75.1 & \underline{75.3} & 88.8 & 83.0 & 85.9 & 82.1 & 82.5 \\
REPLACE & \underline{93.5} & \underline{90.2} & \underline{80.9} & \underline{88.2} & 74.0 & 75.5 & 74.8 & \textbf{90.9} & 88.0 & \underline{89.5} & \underline{84.2} & \underline{84.7} \\
\midrule
CLIP+\textsc{CLoVe} w/o patching & \underline{93.0} & \textbf{91.0} & \textbf{81.6} & \textbf{88.6} & 74.4 & \textbf{77.9} & \textbf{76.1} & 86.2 & \textbf{94.7} & \textbf{90.5} & \textbf{85.1} & \textbf{85.5} \\
CLIP+\textsc{CLoVe} (\(\alpha = .6\)) & \textbf{93.8} & 89.1 & 78.2 & 87.0 & 74.4 & 74.8 & 74.6 & 84.4 & 87.3 & 85.8 & 82.5 & 83.1 \\
\end{tabular}
\caption{Results on SugarCrepe.}
\label{tab:sugarcrepe-results}
\end{table*}

\section{Classification without Prompts}%
\label{sec:classification_without_prompts}

CLIP-like models are evaluated with multiple prompts for classification, typically relying on the ones originally tested by OpenAI's CLIP~\cite{clip}, as we do in this paper.
For example, for ImageNet, there are 80 prompts (templates) used, such as ``a photo of a \{class name\}'' and ``itap of the \{class name\}''.
The reason these prompts are used is that the text representations are usually noisy, and satisfactory average class representation can be obtained from the embeddings for all these texts.
These prompts have been carefully crafted to match the characteristics of the classes and the dataset.
In \cref{tab:common-benchmark-results-without-prompts}, we show the classification results without employing any prompts, just using the class name as the input.
Without patching, our method presents a little drop (2.5\%) in performance regarding the results from \cref{tab:common-benchmark-results}, even when it was tuned to see fully-formed sentences (as opposed to just class names like ``husky'').
When we apply the patching, it drops less in performance in seven out of ten benchmarks than the pre-trained model, and it is on par with 2.

\begin{table*}
\small
\begin{tabular}{c|cccccccccc|cc}
 & \rot{ImageNet} & \rot{Cars} & \rot{CIFAR10} & \rot{CIFAR100} & \rot{MNIST} & \rot{EuroSAT} & \rot{Flowers} & \rot{DTD} & \rot{UCF101} & \rot{HMDB51} & \rot{average} & \rot{average drop} \\
\midrule
pre-trained & \underline{59.0} & \textbf{58.2} & 87.4 & 55.3 & 32.5 & \textbf{48.3} & \textbf{62.4} & 40.5 & 66.9 & 39.2 & 55.0 & 5.1 \\
\midrule
NegCLIP & 54.4 & 45.6 & 85.1 & 57.9 & 31.8 & 30.3 & 51.3 & 37.2 & 64.1 & 38.4 & 49.6 & 3.4 \\
REPLACE & 52.4 & 41.9 & 83.3 & 58.0 & 29.3 & 32.8 & 45.4 & 33.8 & 60.7 & 39.5 & 47.7 & \textbf{2.4} \\
\midrule
CLIP+\textsc{CLoVe} w/o patching & 50.3 & 50.6 & 85.2 & 61.8 & 37.8 & 39.7 & 37.9 & 36.3 & 61.7 & 35.2 & 49.7 & \underline{2.5} \\
CLIP+\textsc{CLoVe} (\(\alpha = .6\)) & \textbf{59.3} & \underline{57.5} & \textbf{88.6} & \textbf{64.6} & \textbf{34.6} & \underline{47.7} & 54.7 & \textbf{43.5} & \textbf{68.0} & \textbf{42.3} & \textbf{56.1} & 4.3 \\
\end{tabular}
\caption{Zero-shot classification results without employing text prompts, which is typically used for CLIP-like models.}
\label{tab:common-benchmark-results-without-prompts}
\end{table*}

\section{Performance in Flickr and COCO Retrieval Tasks}

We evaluate the retrieval performance on Flickr30k~\cite{young-etal-2014-image} and COCO Captions~\cite{coco-captions}, as it is sometimes reported with CLIP-like models~\cite{clip}.
We do not include these results with the main retrieval results because we believe they are near-shot or not zero-shot (in-domain).
NegCLIP and REPLACE fine-tuned on COCO's training set.
Our method is trained on LAION-COCO, whose captions follow a format similar to COCO's.
At the same, COCO images come from Flickr.
We present the results in \cref{tab:common-benchmark-results-retrieval2}.

\begin{table*}
\begin{tabular}{c|cc|cc|c}
 & \multicolumn{2}{c|}{Text-to-Image} & \multicolumn{2}{c|}{Image-to-Text} & \\
 & Flickr30k & COCO Captions & Flickr30k & COCO Captions & avg\onedot{} \\
\midrule
pre-trained & 83.3 & 56.0 & 94.7 & 75.0 & 77.3 \\
\midrule
NegCLIP & \underline{89.5} & 68.5\(^*\) & 95.2 & 79.3\(^*\) & 83.1 \\
REPLACE & \underline{90.0} & \textbf{73.8\(^*\)} & 94.8 & \textbf{83.6\(^*\)} & \textbf{85.6} \\
\midrule
CLIP+\textsc{CLoVe} w/o patching & 87.2 & 65.8 & 87.4 & 68.8 & 77.3 \\
CLIP+\textsc{CLoVe} (\(\alpha = .6\)) & \textbf{90.3} & 68.1 & \textbf{96.3} & 80.0 & 83.7 \\
\end{tabular}
\caption{Retrieval results for Flickr30k and COCO Captions. The evaluation is zero-shot except for those marked with an asterisk (\(^*\)).}
\label{tab:common-benchmark-results-retrieval2}
\end{table*}

\end{document}